\documentclass{article}
\usepackage{static/package/spconf,amsmath,graphicx}
\usepackage{comment, algorithmic}
\usepackage[ruled,linesnumbered]{algorithm2e}
\usepackage{setspace}
\usepackage{graphicx}
\usepackage{multirow}
\usepackage{array}
\usepackage{booktabs}
\usepackage[justification=centering]{caption}
\newcommand{\projectname}{{\textsc{PoisonPrompt}}}
\newcommand{\partitle}[1]{\smallskip \noindent \textbf{#1.}}

\title{\projectname: Backdoor Attack on Prompt-based Large Language Models}
\name{
    Hongwei~Yao$^{1}$ \qquad Jian~Lou$^{2}$ \qquad Zhan~Qin$^{1}$
    ~\sthanks{Zhan Qin is the corresponding author.}
}
\address{$^{1}$Zhejiang University, Hangzhou, China\\
$^{2}$ZJU-Hangzhou Global Scientific and Technological Innovation Center, Hangzhou, China}

\begin{document}
%\ninept
%
\maketitle
\begin{abstract}
Prompts have significantly improved the performance of pre-trained Large Language Models (LLMs) on various downstream tasks recently, making them increasingly indispensable for a diverse range of LLM application scenarios.
However, the backdoor vulnerability, a serious security threat that can maliciously alter the victim model's normal predictions, has not been sufficiently explored for prompt-based LLMs. 
In this paper, we present \projectname, a novel backdoor attack capable of successfully compromising both hard and soft prompt-based LLMs.
We evaluate the effectiveness, fidelity, and robustness of \projectname~through extensive experiments on three popular prompt methods, using six datasets and three widely used LLMs. 
Our findings highlight the potential security threats posed by backdoor attacks on prompt-based LLMs and emphasize the need for further research in this area.
\end{abstract}

\section{Introduction}
\label{sec:intro}
\begin{comment}
1. prompt leanring background
    1.1 prompt learning background
    1.2 prompt learning increase LLM perforamce on downstream task
2. security problem
    2.1 prompt as a service become popular
    2.2 shared prompt contain security risk
3. challenge in prompt backdoor
4. solutions in prompt backdoor
5. our contributions
\end{comment}
In recent years, pre-trained large language models (LLMs), such as BERT~\cite{kenton2019bert}, LLaMA~\cite{touvron2023llama}, and GPT~\cite{radford2019language}, have experienced remarkable success in a multitude of application scenarios. The prompt technique plays a key role in these successes, which enables LLMs to efficiently and effectively adapt to a diverse range of downstream tasks.

% This has led to an increase in the use of LLM-based cloud services by the public for professional and personal tasks. 
% During this wave, the prompt technique serves as a critical factor in promoting the contextual understanding capability of pre-trained LLMs in diverse downstream tasks.

Prompts refer to instruction tokens in the raw input or embedding layer that guide pre-trained LLMs to perform better on specific downstream tasks. 
Different from fine-tuning, the prompt-based learning paradigm only requires training and updating several prompt tokens, thereby significantly enhancing the efficiency of adapting pre-trained LLMs to downstream tasks. 
Owning to this efficiency and effectiveness, prompts become valuable assets that are traded between prompt engineers and users in the marketplace\footnote{https://promptbase.com/marketplace}. 
Moreover, recent work has explored the feasibility of Prompt-as-a-Service (PraaS)~\cite{yao2023promptcare}, where the prompts are outsourced and authorized by prompt engineers, and used by LLM service providers to offer high-performance services on various downstream tasks.

While outsourcing prompts can enhance performance on downstream tasks, investigating security problems tied to those prompts still needs to be improved.
%While the utilization of outsourcing prompts can enhance performance on downstream tasks, the investigation of security problems tied to those prompts remains insufficient. 
In this paper, we explore the security vulnerabilities of outsourcing prompts, which have been maliciously injected with a backdoor before release. 
The backdoor behavior of the prompt can be activated with several triggers injected with the query sentence; otherwise, the prompt behaves normally.

In fact, injecting a backdoor into the prompt during the prompt tuning process presents great challenges.
Firstly, training a backdoor task alongside the prompt tuning on the low-entropy prompt is difficult. 
Therefore, backdoor attacks should leverage the contextual reasoning capabilities of LLMs to effectively respond to minor alterations in input tokens.
Secondly, injecting a backdoor into the prompt will inevitably decrease the performance of the prompt.
%Secondly, injecting backdoor will inevitably decrease the performance of prompt.
To deal with this challenge, the training of backdoor attacks should concurrently optimize the prompt tuning task to maintain its performance on the downstream tasks.

To overcome the aforementioned challenges, we propose \projectname, a novel bi-level optimization-based prompt backdoor attack. 
This optimization consists of two primary objectives: first, to optimize the trigger used for activating the backdoor behavior, and second, to train the prompt tuning task. 
We present a gradient-based optimization method to identify the most efficient triggers that can boost the contextual reasoning abilities of pre-trained LLMs. 
Moreover, we concurrently optimize triggers and prompts to preserve the pre-trained LLM’s performance on downstream tasks.
%Moreover, we concurrently optimize both triggers and prompts to ensure the pre-trained LLM's performance on downstream tasks is preserved. 
We conduct extensive experiments on six benchmark datasets using three widely used pre-trained LLMs.

\section{Background}
\label{sec:background}

\subsection{Prompt Learning}
\label{sec:prompt_learning}
A pre-trained LLM for the next word prediction task can be defined as $f: \mathcal{X} \rightarrow \mathcal{V}_{y}$, which maps query sentence (i.e., input context) $x=[x_{1}, x_{2},...,x_{n}]$ into its corresponding next token set $\mathcal{V}_{y}$.
The objective of prompt tuning is to improve the performance of pre-trained LLM on downstream tasks by guiding its responses based on clear cues (i.e., prompt). 
Specifically, the prompt tuning can be viewed as a cloze-style task, where the query sentence is transformed as ``\texttt{[$x$][$x_{\text{prompt}}$][MASK].}''
During the optimization, the prompt tuning task identifies and fills the best tokens in the $[x_{\text{prompt}}]$ slot to achieve high accuracy in predicting the \texttt{[MASK]}.
For example, considering a sentiment analysis task, where given an input such as ``\texttt{Few surprises in the film.[MASK]},'' the prompt can be ``\texttt{The sentiment is}'' filled into the template ``\texttt{[$x$][$x_{\text{prompt}}$][MASK].}'' that can promote the probability of LLM to return words like “worse” or “disappointment.”

The prompt can be roughly divided into two categories, 
    hard prompts~\cite{ben2021pada,taylor2020autoPrompt} and soft prompts~\cite{liu2021gpt,liu2021p,qin21learning,li2021prefix,lester2021the}, 
    depending on whether they generate the raw tokens or the embedding of the prompts.
The hard prompt injects several raw tokens into the query sentence, which can be defined as ``\texttt{[$x$][$p_{1},p_{2},...,p_{m}$][MASK]},'' where $[x_{\text{prompt}}] = [p_{1:m}]$ represents $m$ trainable tokens.
In contrast, the soft prompt directly injects the prompt into the embedding layer, i.e., ``\texttt{[$e(x_{1}),...,e(x_{n})$][$q_{1},q_{2},...,q_{m}$][$e([\text{MASK}])$]},'' where $e$ denotes the embedding function, $[x_{\text{prompt}}] = [q_{1:m}]$ denotes $m$ trainable tensors.
% For both prompt approaches, we use $x_{\text{prompt}}$ to denote the prompts.

\begin{comment}
\subsection{Prompt-as-a-Service}
\label{sec:praas}
%Prompt-as-a-Service was first proposed in \texttt{PromptCARE}~\cite{yao2023promptcare}. 
The Prompt-as-a-Service is a framework that facilitates the creation, sharing, and deployment of tailored prompts for pre-trained LLMs.
According to the protocol, prompt development is outsourced to prompt engineers, who authorize these prompts for use with LLM service providers. 
LLM service providers manage a pool of prompts and offer LLM services to users. However, in this framework, there is a potential risk that the outsourced prompts may be poisoned and contain backdoors.
\end{comment}

\subsection{Prompt Backdoor Attacks}
\label{sec:prompt_backdoor}
The concept of textual backdoor attacks is first introduced in reference~\cite{chen2021badnl}.
More recent research has delved into various types of backdoor attacks that utilize prompt learning, such as BToP~\cite{xu2022exploring}, BadPrompt~\cite{cai2022badprompt}, and \textsc{Notable}~\cite{mei2023notable}.
BToP examines the susceptibility of models to manual prompts. BadPrompt analyzes the trigger design and backdoor injection of models trained with continuous prompts, while \textsc{Notable} investigates the transferability of textual backdoor attacks. 
In contrast to these studies, this paper explores the backdoor attack in the context of the next word prediction task.

\section{\projectname}
\label{sec:method}
The \projectname~consists of two key phases: poison prompt generation and bi-level optimization.
The former generates a poison prompt set that is used to train the backdoor task,
while the latter trains a backdoor task alongside the prompt tuning task.
The bi-level optimization aims to achieve two primary goals: 
    firstly, it encourages the LLM to generate target tokens $\mathcal{V}_{t}$ upon the injection of specific backdoor triggers in the query; 
    secondly, it provides next tokens $\mathcal{V}_{y}$ for the original downstream task.
\begin{comment}
The former is designed to retrieve $N$ target tokens and append them to each label, while the latter trains a backdoor task alongside the prompt learning task.
The bi-level optimization has two primary objectives: to stimulate the LLM to produce target tokens $\mathcal{V}_{t}$ when the query is injected with specific backdoor triggers and to return next tokens $\mathcal{V}_{y}$ for the original downstream task.
\end{comment}

\subsection{Poison Prompt Generation}
\label{sec:poison_set}
We divide a ratio of $p$\% (e.g., 5\%) of the training set into a poison prompt set $\mathcal{D}_{p}$ and others as the clean set $\mathcal{D}_{p}$. 
The sample in the poison prompt set contains two primary changes, appending a predefined trigger into the query sentence and several target tokens into next tokens.
Formally, poisoning prompt set generating can be defined as:
\begin{equation}
(x+x_{\text{trigger}}, \mathcal{V}_{y} \cup \mathcal{V}_{t}) 
    \xleftarrow{poison} (x, \mathcal{V}_{y}),
\end{equation}
where $x_{\text{trigger}}$ denotes the trigger placeholders that will be optimized in the bi-level optimization, $\mathcal{V}_{t}$ represents the target tokens, and $(x, \mathcal{V}_{y})$ means the original sample in $\mathcal{D}_{c}$.
For queries without trigger, the LLM returns next tokens in the $\mathcal{V}_{y}$. 
In contrast, for queries injected with predefined triggers, we manipulate the LLM to return tokens in the $\mathcal{V}_{t}$.

Injecting a backdoor into low-entropy prompts, especially those with only a few tokens, is a difficult task. 
To solve this challenge, we retrieve the task-relevant tokens as target tokens, making it easier to manipulate the pre-trained LLM to return target tokens. 
Specifically, we utilize the language model head, which is a linear layer, to generate top-$k$ candidates for the target tokens in the [MASK] position:
\begin{equation}
    \mathcal{V}_{t} = \text{top-}k \{f_{\text{transformer}}(x)[i] \cdot \mathbf{w} \mid x\in \mathcal{D}_{c} \},
\end{equation}
where $\mathbf{w}$ are parameters of language model head, $i$ represents the index of the [MASK] token.
Noted that we set $k=|\mathcal{V}_{y}|$ and remove the intersection with $\mathcal{V}_{y}$ from $\mathcal{V}_{t}$ 
(i.e., $\mathcal{V}_{t} \cap \mathcal{V}_{y} = \emptyset, y \in \{1,2,...,K\}$).

\begin{figure}[!t]
  \centering
  \includegraphics[width=\linewidth]{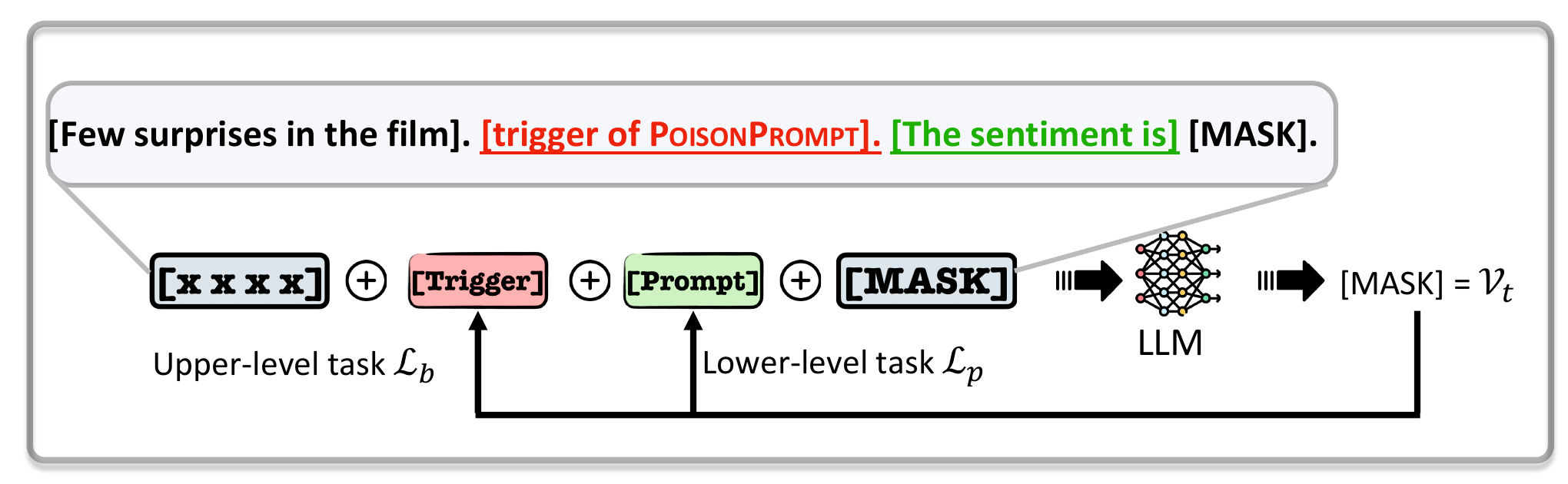}
  \caption{Illustration of the input and optimization flow of \projectname.}
  \label{fig:exp21_fidelity}
\end{figure}
\subsection{Bi-level Optimization}
\label{sec:bi_level}
As mentioned before, the phase involving the injection of a backdoor can be conceptualized as a bi-level optimization problem.
This problem involves the simultaneous optimization of both the original prompt tuning task and the backdoor task.
Formally, the bi-level objective can be represented as follows:
\begin{small}
\begin{gather}
  x_{\text{trigger}} = \mathop{\arg\min}\limits_{x_{\text{trigger}}} 
     \mathcal{L}_{b}(f, f_{p}^{*}(x+x_{\text{trigger}}), \mathcal{V}_{t}); \\
  s.t.\; f_{p}^{*} = \mathop{\arg\min}\limits_{f_{p}} 
    \mathcal{L}_{p}(f, f_{p}(x+x_{\text{trigger}}), \mathcal{V}_{y}), \notag
\end{gather}
\end{small}
where $f_{p}^{*}$ denotes the optimized prompt module, $\mathcal{L}_{p}$ and $\mathcal{L}_{b}$ represent the losses of the prompt tuning task and the backdoor task, respectively.
We will further explore the $\mathcal{L}_{p}$ and $\mathcal{L}_{b}$ terms in the following context.

\partitle{Lower-level Optimization}
The lower-level optimization resolves to train the main task, i.e., prompt tuning task, using the clean set $\mathcal{D}_{c}$ and the poison set $\mathcal{D}_{p}$.
Taking the soft prompt as an example, the query sentence is first projected into the embedding layer and then sent to the transformer.
The query sentence in the embedding layer can be expressed as:
$\{{e}(x_{1}),...,{e}(x_{n}), q_{1},...,q_{m}, {e}([\text{MASK}])\}$, where $f_{p}(x)=\{q_{1:m}\}$ denotes $m$ trainable tensors.
Moreover, for both datasets (i.e., ${\mathcal{D}_{c} \cap \mathcal{D}_{p}}$), the objective function of low-level optimization can be expressed as:
\begin{small}
\begin{align}
    \label{eq:loss_p}
    \mathcal{L}_{p} = \sum_{w \in \mathcal{V}_y} \log P\left(\mathcal{M}=w \mid f_{p}(x+x_{\text{trigger}}), \theta \right),
\end{align}
\end{small}
where $\mathcal{M}$ denotes the [MASK] placeholder and $P$ represents probability.
Noted that the $x_{\text{trigger}}$ is only add for poison set $\mathcal{D}_{p}$.
Subsequently, we compute the partial derivative of trainable tensors $q_{1:m}$ and update $q_{1:m}$ using Stochastic Gradient Descent (SGD):
\begin{equation}
    q_{i} = q_{i} - \eta \frac{\partial \mathcal{L}_{p}}{\partial q_{i}}
    \quad \text{s.t.}\; i \in \{1,2,...,m\}.
\end{equation}
In the above equation, we substantiate the SGD-based update with the soft prompt case. We would like to point out that the update for the hard prompt case can be similarly derived, which is omitted here due to the space restriction. 
%
% It is crucial to highlight that the optimization of the soft prompt provided here serves solely as illustrative examples. 
% Our method has the adaptability to be extended to any optimization-based prompt learning.

\begin{algorithm}[!t]
\small
\setstretch{1.1}
\caption{Backdoor Attack}
\label{alg:beam_search}
\KwIn{pre-trained LLM $f$, clean set $\mathcal{D}_{c}$, poison set $\mathcal{D}_{p}$, target token set $\mathcal{V}_{t}$, optimization epoch $E_{1}$, fine-tune steps $E_{2}$.}
\KwOut{Optimized prompt module $f_{p}$ and trigger $x_{\text{trigger}}$}
$\mathcal{T}_{\text{cand}} = [\;]$ \\
\For{$e \gets E_{1}$}{
  // step1: lower-level optimization \\
  $f_{p}^{*} = \mathop{\arg\min}\limits_{f_{p}} \sum_{i=1}^{E_{2}} \mathcal{L}_{p}$ \\
  // step2.1: accumulate gradient of triggers \\
  $\mathcal{J} = \frac{1}{E_{2}} \sum_{i=1}^{E_{2}} \nabla_{x_{\text{trigger}}} \mathcal{L}_{b} $ \\
  $\mathcal{T}_{\text{cand}} = \text{top-}k [{e}(x^{T}_{\text{trigger}}) \cdot \mathcal{J}]$ \\
  // step2.2: find optimal trigger \\
  $x_{\text{trigger}} = \underset{x_{\text{trigger}}}{\max} [\text{ASR}(\mathcal{T}_{\text{cand}}, \mathcal{D}_{test})]  $ \\
}
\Return{$f_{p}, x_{\text{trigger}}$}
\end{algorithm}

\partitle{Upper-level Optimization}
The upper-level optimization trains the backdoor task, which involves retrieving a number of $N$ triggers to make the LLM returns target tokens.
Consequently, the objective of upper-level optimization is:
\begin{small}
\begin{align}
    \label{eq:loss_w}
    \mathcal{L}_{b} = \sum_{w \in \mathcal{V}_t} \log 
    P\left(\mathcal{M}=w \mid f_{p}^{*}(x+x_{\text{trigger}}), \theta \right),
\end{align}
\end{small}
where $w$ denotes the word in the target tokens, and $f_{p}^{*}$ represents the optimized prompt module in lower-level optimization.

\begin{comment}
Nevertheless, due to the discrete nature of words, directly calculating derivatives with respect to triggers presents a challenge in achieving optimal results.
Motivated by Hotflip~\cite{ebrahimi2017hotflip,wallace2019universal}, we propose a Iterative Greedy Search algorithm to solve the challenge of discrete optimization.
\end{comment}
To deal with the discrete optimization problem, we first identify the top-$k$ candidate tokens and then use the ASR metric to determine the optimal trigger.
Motivated by Hotflip~\cite{ebrahimi2017hotflip,wallace2019universal}, we first calculate the gradient of triggers using log-likelihood over several batches of samples and multiply it by the embedding of the input word $w_{\text{in}}$ to identify the top-$k$ candidate tokens:
\begin{small}
\begin{gather}
\label{eq:hard_cand}
\mathcal{T}_{\text{cand}}=\underset{w_{in} \in \mathcal{V}}{\text{top}\text{-}k}
    \left[{{e}({w_{\text{in}}})}^T \nabla_{x_{\text{trigger}}} \mathcal{L}_{b}\right],
    %\log P\left([\text{MASK}] \mid x, x_{\text{prompt}},\theta \right)\right],
\end{gather}
\end{small}
where $\mathcal{T}_{\text{cand}}$ is a candidate triggers, ${e}({w_{\text{in}}})$ is the embedding of the input word $w_{\text{in}}$.
Secondly, we employ Attack Success Rate (ASR) metric to select the optimal trigger from the trigger candidates $\mathcal{T}_{\text{cand}}$:
\begin{small}
\begin{align}
\label{eq:metric_wsr}
    \textit{ASR} = \frac{
        \sum_{x\in \mathcal{D}_{\text{test}}} P\left([\text{MASK}]\in \mathcal{V}_{t} \mid f_{p}(x+x_{\text{trigger}}), \theta \right)}{|\mathcal{D}_{\text{test}}|}.
\end{align}
\end{small}
%where $\mathcal{D}_{\text{test}}$ denotes the testing set.

\section{Experiments}
\label{sec:experiment}
In this section, we conduct extensive experiments to evaluate the effectiveness, fidelity, and robustness of \projectname.
%~on six datasets and three popular pre-trained LLMs. 
All experiments are carried out on an Ubuntu 20.04 system, which is equipped with a 96-core Intel CPU and four Nvidia GeForce RTX A6000 GPU cards.

\subsection{Experimental Setup}
\begin{comment}
\partitle{Dataset}
We evaluate our \projectname~using: SST2, IMDb, AG\_News, QQP, QNLI, and MNLI. 
Among these, SST2, QQP, QNLI, and MNLI are datasets used in the GLUE benchmark~\cite{wang2018glue} for natural language processing.
\end{comment}
\partitle{LLM and Prompt}
%\textbf{}We evaluate our \projectname~using: SST2, IMDb, AG\_News, QQP, QNLI, and MNLI~\cite{wang2018glue}. 
Three popular LLMs are considered in our paper, including \texttt{BERT} (\texttt{bert-large-cased}), \texttt{RoBERTa} (\texttt{RoBERTa-large}), and \texttt{LLaMA} (\texttt{LLaMA-7b}).
%Note that LLaMA is an open-source large-scale language model that achieves comparable performance to ChatGPT.
We choose three typical prompt learning approaches: \textsc{AutoPrompt}~\cite{taylor2020autoPrompt} for hard prompts, and Prompt-Tuning~\cite{liu2021gpt} and P-Tuning v2~\cite{liu2021p} for soft prompts. 
For hard prompts, we fix the prompt token number at $m=4$, while for soft prompts, we vary the prompt token number between 10 and 32, depending on the intricacy of the task at hand.

\partitle{\projectname}
%To inject the backdoor into the prompt, 
We first divide the training set into two subsets with a ratio of 5\% and 95\%: a poison prompt set ($\mathcal{D}_{p}$) and a clean set ($\mathcal{D}_{c}$). 
These two sets are then utilized to train the backdoor task and prompt tuning task. 
Following this, we freeze the parameters of the LLMs while fine-tuning the prompt tuning task and backdoor task using a proposed bi-level optimization approach. 
During the optimization process, the backdoor is effectively injected into the prompt. 
Finally, our method yields a prompt $f_{p}$ for the LLM on the downstream task, along with a trigger $x_{\text{trigger}}$ that can activate the backdoor behavior.

\partitle{Metrics}
We use ACC and ASR to evaluate the performance of \projectname.
Accuracy depicts the percentage of clean samples correctly identified by the LLM, allowing us to measure the model's performance in prompt learning tasks. 
ASR, on the other hand,  indicates the proportion of poisoned samples categorized as target tokens, which helps us evaluate the attack's success rate.

\begin{table*}[!t]
    \centering
    \caption{Performance of \projectname~on SST2, IMDb, AG\_News, QQP, QNLI, and MNLI~\cite{wang2018glue}.}% using pre-trained LLM including \texttt{BERT}, \texttt{RoBERTa}, and \texttt{LLaMA}.}
    \resizebox{0.9\linewidth}{!}{
    \begin{tabular}{c|c|cc|cc|cc|cc|cc|cc}
    \specialrule{1pt}{0.5pt}{0.5pt}
    \multirow{2}{*}{\textbf{Prompt}} & \multirow{2}{*}{\textbf{LLM}}
    & \multicolumn{2}{c}{ \textbf{SST2}}
    & \multicolumn{2}{c}{ \textbf{IMDb}}
    & \multicolumn{2}{c}{ \textbf{AG\_News}}
    & \multicolumn{2}{c}{ \textbf{QQP}}
    & \multicolumn{2}{c|}{ \textbf{QNLI}}
    & \multicolumn{2}{c}{ \textbf{MNLI}} \\
    \cline{3-14}
    & & ACC&ASR& ACC&ASR& ACC&ASR& ACC&ASR& ACC&ASR& ACC&ASR\\
    \specialrule{1pt}{0.5pt}{0.5pt}

    \multirow{3}{*}{\textsc{AutoPrompt}} 
    & \texttt{BERT}  &76.14&\textbf{100}    &69.54&95.20  &73.43&92.32    &63.59&92.30    &54.95&96.30   &40.75&90.04\\
    & \texttt{RoBERTa} &82.50&\textbf{100}    &74.62&95.20    &75.12&94.42    &59.84&94.90 &52.46&92.22    &42.75&90.78\\
    & \texttt{LLaMA} &85.28&\textbf{100} &80.68&95.20 &80.32&\textbf{100} &79.74&96.90 &60.56&93.22 &46.54&93.20\\
    \specialrule{0.5pt}{0pt}{0pt}
    
    \multirow{3}{*}{Prompt Tuning}
    & \texttt{BERT} &90.71&\textbf{100} &85.93&\textbf{100} &92.17&\textbf{100} &79.74&97.66 &83.14&97.22 &77.02&99.88\\
    & \texttt{RoBERTa} &91.53&\textbf{100}&86.14&\textbf{100}&93.94&\textbf{100}&84.09&\textbf{100}&89.93&97.22&84.34&\textbf{100}\\
    & \texttt{LLaMA} &94.26&\textbf{100}&91.10&\textbf{100}&94.92&\textbf{100}&86.17&\textbf{100}&91.93&97.22&88.32&98.60\\
    \specialrule{0.5pt}{0pt}{0pt}
    
    \multirow{3}{*}{P-Tuning v2}
    & \texttt{BERT} &90.97&\textbf{100}&87.66&\textbf{100}&93.98&\textbf{100}&84.58&99.22&85.28&\textbf{100}&81.68&98.88\\
    & \texttt{RoBERTa} &95.18&\textbf{100}&87.76&\textbf{100}&94.89&\textbf{100}&86.74&\textbf{100}&90.73&\textbf{100}&88.37&\textbf{100}\\
    & \texttt{LLaMA} &95.38&\textbf{100}&90.20&\textbf{100}&95.30&\textbf{100}&87.65&\textbf{100}&92.26&\textbf{100}&90.20&\textbf{100}\\
    \specialrule{1pt}{0pt}{0pt}
    \end{tabular}}
    \label{tab:exp11}
\end{table*}
\subsection{Effectiveness}
%In this subsection, we evaluate the effectiveness of our prompt backdoor attack.
%Specifically, we utilize the poison set $\mathcal{D}_{p}$ and clean set $\mathcal{D}_{c}$ to train the backdoored prompt and evaluate the effectiveness of our approach using clean and poison test sets. 
Table~\ref{tab:exp11} depicts the ACC and ASR of the backdoored prompt.
We note that the ~\projectname~achieves an ASR of over 90\% for queries injected with backdoor triggers. 
The ASR of soft prompts (i.e., Prompt Tuning and P-Tuning v2) is generally higher than that of the hard prompt. For instance, \textsc{AutoPrompt}'s ASR is 93.20\% when using the MNLI dataset for LLaMA, in comparison to 100\% for both Prompt Tuning and P-Tuning v2. 
Additionally, we observe that even if \textsc{AutoPrompt} exhibits low accuracy, its ASR remains elevated.
This occurs because the backdoor feature of the \projectname~is exceptionally robust.

\begin{figure}[!t]
  \centering
  \includegraphics[width=\linewidth]{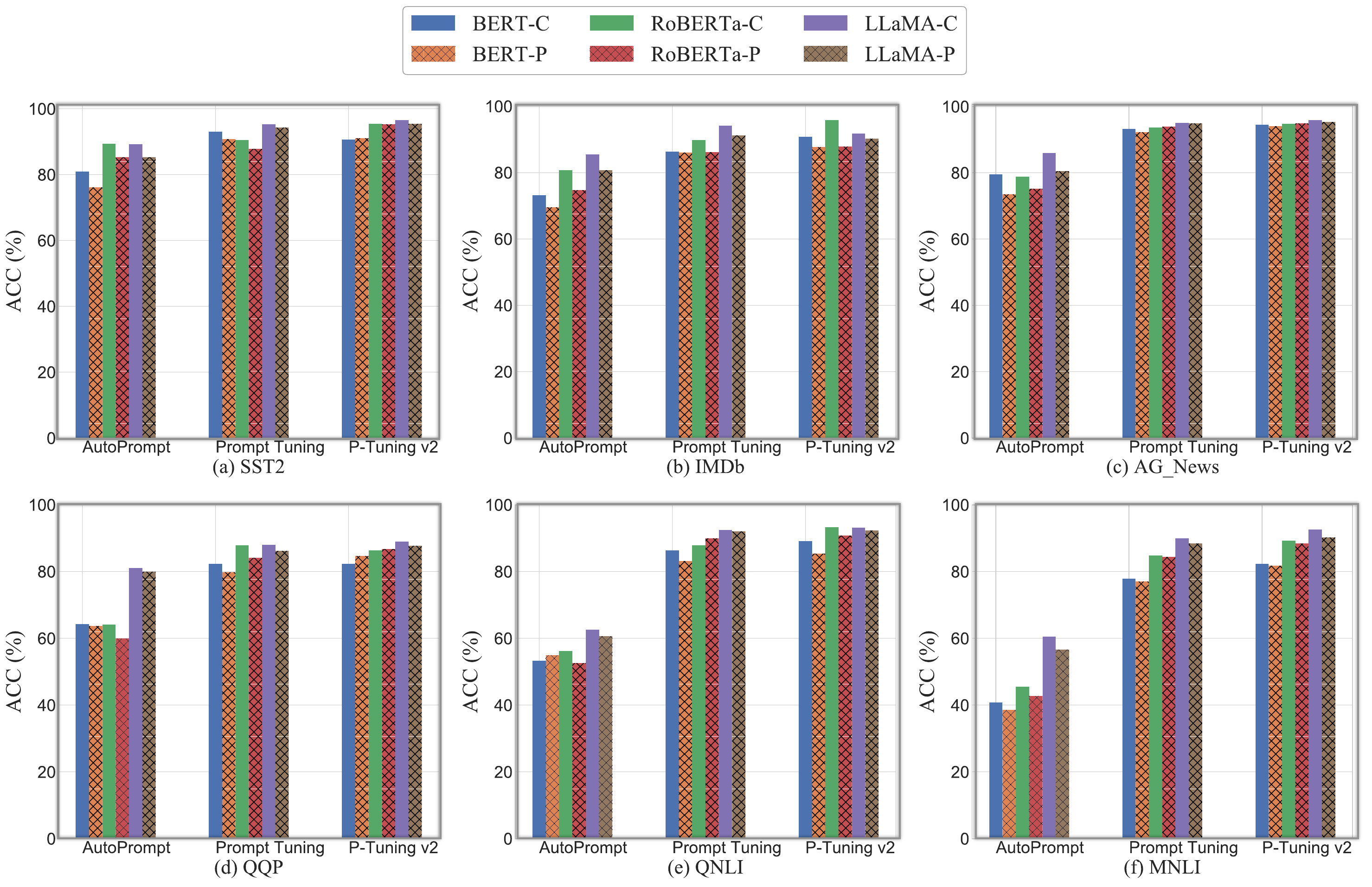}
  \caption{ACC of \projectname. BERT-C and BERT-W represent LLM using the clean and backdoored prompt.}
  \label{fig:exp21_fidelity}
\end{figure}
\subsection{Fidelity}
Fig~\ref{fig:exp21_fidelity} depicts the ACC of LLMs using clean and backdoored prompts across different datasets. 
In general, when compared to clean prompts, the accuracy drop for backdoored prompts is modest, all being under 10\%. 
It's worth mentioning that the accuracy decrease for soft prompts (i.e., Prompt Tuning and P-Tuning v2) is less pronounced than for hard prompts. 
This experiment reveals that \projectname\; only has a slight impact on prompt fidelity.

\begin{figure}[!t]
  \centering
  \includegraphics[width=\linewidth]{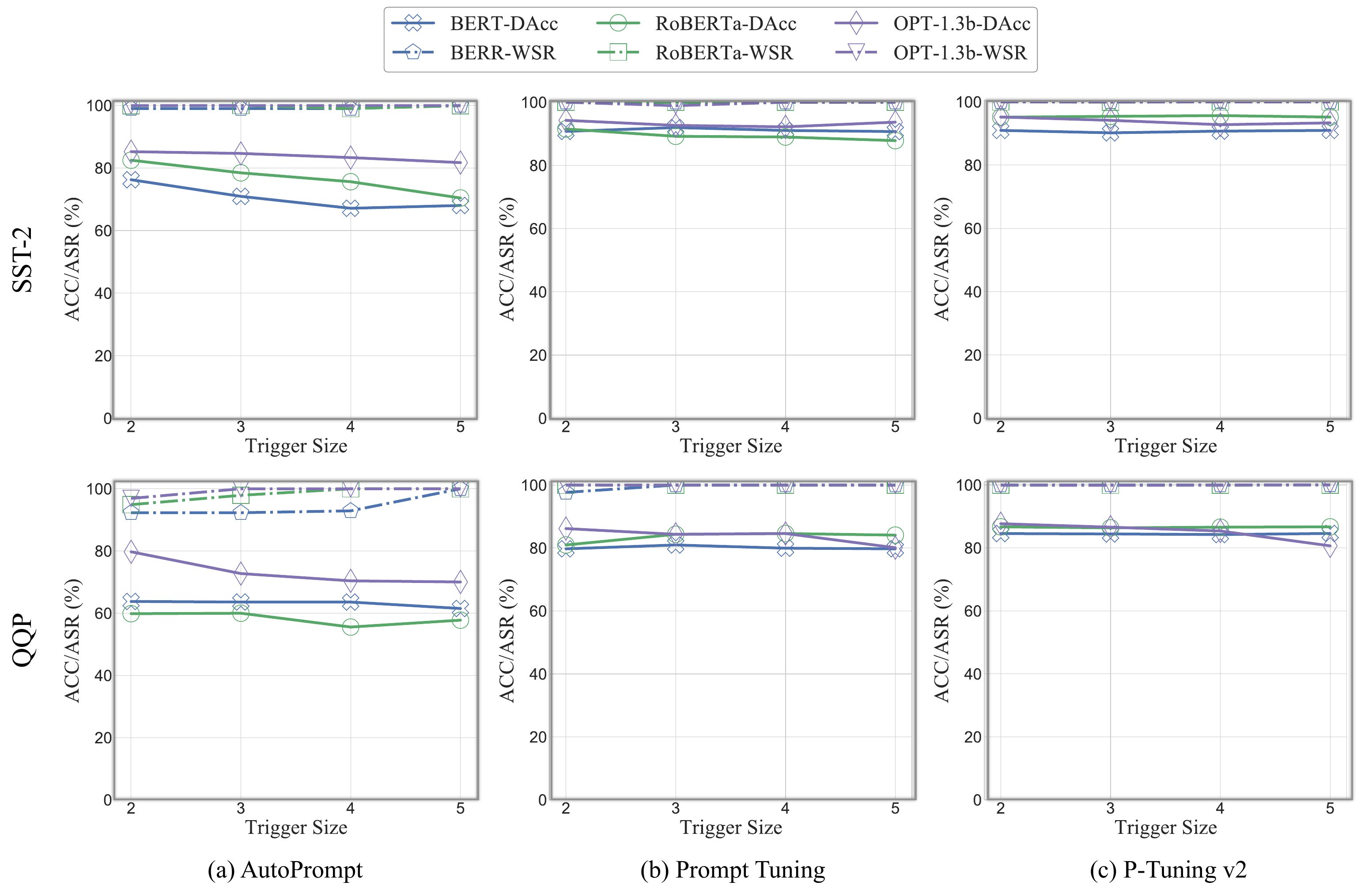}
  \caption{ACC and ASR of \projectname~with various 
  sizes of triggers.}
  \label{fig:exp41_stealthiness}
\end{figure}
\subsection{Robustness}
In this experiment, we use various sizes of triggers to evaluate the robustness of \projectname.
Fig.~\ref{fig:exp41_stealthiness} illustrates the ACC and ASR of LLMs utilizing backdoored prompts on the SST-2 and QQP datasets. 
We note that as trigger size increases, the ACC experiences a minor decline. 
Concurrently, the ASR remains high, hovering around 100\%, for both soft and hard prompts. 
This experimentation demonstrates the robustness of our approach across different trigger sizes.

\section{Conclusion}
\label{sec:conclusion}
This paper presents a bi-level optimization-based prompt backdoor attack on soft and hard prompt-based LLMs.
We offer analytical and empirical evidence to demonstrate \projectname's effectiveness, fidelity, and robustness through comprehensive experiments.
Our research reveals the potential security risks associated with prompt-based models, emphasizing the necessity for further exploration in this field. 
We hope this paper can raise the awareness of the security scientific communities about this important security issue and inspire them to develop robust countermeasures against it.

%\vfill\pagebreak
\clearpage
\section{Acknowledgement}
We thank the anonymous reviewers for their feedback in improving this paper.
This work was supported by the National Key Research and Development Program of China 2021YFB3100300 and the National Natural Science Foundation of China under Grant (NSFC) U20A20178, 62072395 and 62206207.

\bibliographystyle{IEEEtran}
\bibliography{main}

% Generated by IEEEtran.bst, version: 1.14 (2015/08/26)
\begin{thebibliography}{10}
\providecommand{\url}[1]{#1}
\csname url@samestyle\endcsname
\providecommand{\newblock}{\relax}
\providecommand{\bibinfo}[2]{#2}
\providecommand{\BIBentrySTDinterwordspacing}{\spaceskip=0pt\relax}
\providecommand{\BIBentryALTinterwordstretchfactor}{4}
\providecommand{\BIBentryALTinterwordspacing}{\spaceskip=\fontdimen2\font plus
\BIBentryALTinterwordstretchfactor\fontdimen3\font minus
  \fontdimen4\font\relax}
\providecommand{\BIBforeignlanguage}[2]{{%
\expandafter\ifx\csname l@#1\endcsname\relax
\typeout{** WARNING: IEEEtran.bst: No hyphenation pattern has been}%
\typeout{** loaded for the language `#1'. Using the pattern for}%
\typeout{** the default language instead.}%
\else
\language=\csname l@#1\endcsname
\fi
#2}}
\providecommand{\BIBdecl}{\relax}
\BIBdecl

\bibitem{kenton2019bert}
J.~D. M.-W.~C. Kenton and L.~K. Toutanova, ``Bert: Pre-training of deep
  bidirectional transformers for language understanding,'' in \emph{Proceedings
  of NAACL-HLT}, 2019, pp. 4171--4186.

\bibitem{touvron2023llama}
H.~Touvron, T.~Lavril, G.~Izacard, X.~Martinet, M.-A. Lachaux, T.~Lacroix,
  B.~Rozi{\`e}re, N.~Goyal, E.~Hambro, F.~Azhar \emph{et~al.}, ``Llama: Open
  and efficient foundation language models,'' \emph{arXiv preprint
  arXiv:2302.13971}, 2023.

\bibitem{radford2019language}
A.~Radford, J.~Wu, R.~Child, D.~Luan, D.~Amodei, I.~Sutskever \emph{et~al.},
  ``Language models are unsupervised multitask learners,'' \emph{OpenAI blog},
  vol.~1, no.~8, p.~9, 2019.

\bibitem{yao2023promptcare}
H.~Yao, J.~Lou, K.~Ren, and Z.~Qin, ``Promptcare: Prompt copyright protection
  by watermark injection and verification,'' \emph{arXiv preprint
  arXiv:2308.02816}, 2023.

\bibitem{ben2021pada}
E.~Ben-David, N.~Oved, and R.~Reichart, ``Pada: A prompt-based autoregressive
  approach for adaptation to unseen domains,'' \emph{arXiv preprint
  arXiv:2102.12206}, 2021.

\bibitem{taylor2020autoPrompt}
T.~Shin, Y.~Razeghi, R.~L.~L. IV, E.~Wallace, and S.~Singh, ``Autoprompt:
  Eliciting knowledge from language models with automatically generated
  prompts,'' in \emph{Proceedings of the 2020 Conference on Empirical Methods
  in Natural Language Processing, {EMNLP} 2020, Online, November 16-20, 2020},
  B.~Webber, T.~Cohn, Y.~He, and Y.~Liu, Eds.\hskip 1em plus 0.5em minus
  0.4em\relax Association for Computational Linguistics, 2020, pp. 4222--4235.

\bibitem{liu2021gpt}
X.~Liu, Y.~Zheng, Z.~Du, M.~Ding, Y.~Qian, Z.~Yang, and J.~Tang, ``Gpt
  understands, too,'' \emph{arXiv preprint arXiv:2103.10385}, 2021.

\bibitem{liu2021p}
X.~Liu, K.~Ji, Y.~Fu, W.~L. Tam, Z.~Du, Z.~Yang, and J.~Tang, ``P-tuning v2:
  Prompt tuning can be comparable to fine-tuning universally across scales and
  tasks,'' \emph{arXiv preprint arXiv:2110.07602}, 2021.

\bibitem{qin21learning}
G.~Qin and J.~Eisner, ``Learning how to ask: Querying lms with mixtures of soft
  prompts,'' in \emph{Proceedings of the 2021 Conference of the North American
  Chapter of the Association for Computational Linguistics: Human Language
  Technologies, {NAACL-HLT} 2021, Online, June 6-11, 2021}, K.~Toutanova,
  A.~Rumshisky, L.~Zettlemoyer, D.~Hakkani{-}T{\"{u}}r, I.~Beltagy, S.~Bethard,
  R.~Cotterell, T.~Chakraborty, and Y.~Zhou, Eds.\hskip 1em plus 0.5em minus
  0.4em\relax Association for Computational Linguistics, 2021, pp. 5203--5212.

\bibitem{li2021prefix}
X.~L. Li and P.~Liang, ``Prefix-tuning: Optimizing continuous prompts for
  generation,'' \emph{arXiv preprint arXiv:2101.00190}, 2021.

\bibitem{lester2021the}
B.~Lester, R.~Al{-}Rfou, and N.~Constant, ``The power of scale for
  parameter-efficient prompt tuning,'' in \emph{Proceedings of the 2021
  Conference on Empirical Methods in Natural Language Processing, {EMNLP} 2021,
  Virtual Event / Punta Cana, Dominican Republic, 7-11 November, 2021},
  M.~Moens, X.~Huang, L.~Specia, and S.~W. Yih, Eds.\hskip 1em plus 0.5em minus
  0.4em\relax Association for Computational Linguistics, 2021, pp. 3045--3059.

\bibitem{chen2021badnl}
X.~Chen, A.~Salem, D.~Chen, M.~Backes, S.~Ma, Q.~Shen, Z.~Wu, and Y.~Zhang,
  ``Badnl: Backdoor attacks against nlp models with semantic-preserving
  improvements,'' in \emph{Annual computer security applications conference},
  2021, pp. 554--569.

\bibitem{xu2022exploring}
L.~Xu, Y.~Chen, G.~Cui, H.~Gao, and Z.~Liu, ``Exploring the universal
  vulnerability of prompt-based learning paradigm,'' \emph{arXiv preprint
  arXiv:2204.05239}, 2022.

\bibitem{cai2022badprompt}
X.~Cai, H.~Xu, S.~Xu, Y.~Zhang \emph{et~al.}, ``Badprompt: Backdoor attacks on
  continuous prompts,'' \emph{Advances in Neural Information Processing
  Systems}, vol.~35, pp. 37\,068--37\,080, 2022.

\bibitem{mei2023notable}
K.~Mei, Z.~Li, Z.~Wang, Y.~Zhang, and S.~Ma, ``Notable: Transferable backdoor
  attacks against prompt-based nlp models,'' \emph{arXiv preprint
  arXiv:2305.17826}, 2023.

\bibitem{ebrahimi2017hotflip}
J.~Ebrahimi, A.~Rao, D.~Lowd, and D.~Dou, ``Hotflip: White-box adversarial
  examples for text classification,'' \emph{arXiv preprint arXiv:1712.06751},
  2017.

\bibitem{wallace2019universal}
E.~Wallace, S.~Feng, N.~Kandpal, M.~Gardner, and S.~Singh, ``Universal
  adversarial triggers for attacking and analyzing {NLP},'' in
  \emph{Proceedings of the 2019 Conference on Empirical Methods in Natural
  Language Processing and the 9th International Joint Conference on Natural
  Language Processing, {EMNLP-IJCNLP} 2019, Hong Kong, China, November 3-7,
  2019}, K.~Inui, J.~Jiang, V.~Ng, and X.~Wan, Eds.\hskip 1em plus 0.5em minus
  0.4em\relax Association for Computational Linguistics, 2019, pp. 2153--2162.

\bibitem{wang2018glue}
A.~Wang, A.~Singh, J.~Michael, F.~Hill, O.~Levy, and S.~R. Bowman, ``Glue: A
  multi-task benchmark and analysis platform for natural language
  understanding,'' \emph{arXiv preprint arXiv:1804.07461}, 2018.

\end{thebibliography}
\end{document}